\documentclass[10pt,twocolumn,letterpaper]{article}

\usepackage{iccv}
\usepackage{times}
\usepackage{epsfig}
\usepackage{graphicx}
\usepackage{amsmath}
\usepackage{amssymb}
\usepackage{amsfonts}
\usepackage{amssymb}
\usepackage{dsfont}
\usepackage{mathtools}
\usepackage{enumitem}
\usepackage{bm}
\usepackage{appendix}

\def\@onedot{\ifx\@let@token.\else.\null\fi\xspace}
\def\eg{\emph{e.g}\onedot} 
\def\ie{\emph{i.e}\onedot} 
 
 \def\vs{\emph{vs}\onedot} 
\def\wrt{w.r.t\onedot}

\def\fig#1{Figure~\ref{fig:#1}}
\def\tab#1{Table~\ref{tab:#1}}
\def\sect#1{Section~\ref{sec:#1}}
\def\Eq#1{Eq.~(\ref{eq:#1})}
\def\Ind#1{[\![#1]\!]}  %indicator function

\def\mypar#1{\vspace{1mm}{\noindent\bf #1.}\hspace{1mm}}

\def\figvspaceOne{\vspace{-4mm}}
\def\figvspaceTwo{\vspace{-2mm}}

\usepackage[breaklinks=true,bookmarks=false,colorlinks=True]{hyperref}

\iccvfinalcopy % *** Uncomment this line for the final submission

\ificcvfinal\fi

\begin{document}
%%%%%%%%% TITLE
\title{Adaptative Inference Cost With  
Convolutional Neural Mixture Models}

\author{Adria Ruiz  \hspace{1cm} Jakob Verbeek\\
Univ.\ Grenoble Alpes, Inria, %\\
CNRS, Grenoble INP, LJK, 38000 Grenoble, France\\
{\tt\small firstname.lastname@inria.fr}
}

\maketitle
\ificcvfinal\thispagestyle{empty}\fi

%%%%%%%%% ABSTRACT
\begin{abstract}
Despite the outstanding performance of convolutional neural networks (CNNs) for many vision tasks, 
the required computational cost during inference is problematic when resources are limited. In this context, we propose Convolutional Neural Mixture Models (CNMMs), a probabilistic model embedding a large number of CNNs that can be jointly trained and evaluated in an efficient manner. 
Within the proposed framework, we present different mechanisms to prune subsets of CNNs from the mixture, allowing to easily adapt the computational cost required for inference. 
Image classification and semantic segmentation experiments show that our method  achieve excellent  accuracy-compute trade-offs. 
Moreover, unlike most of previous approaches, a single CNMM provides a large range of operating points along this trade-off, without any re-training.

\end{abstract}

\section{Introduction}
\label{sec:introduction}
Convolutional neural networks (CNNs) form the basis of many state-of-the-art  computer vision models.
Despite their outstanding performance, the computational cost of inference in these CNN-based models is typically very high. 
This holds back applications on mobile platforms, such as autonomous vehicles, drones, or  phones, 
where computational resources are limited, 
concurrent data-streams need to be processed, and low-latency prediction is critical. 

\begin{figure}
  \includegraphics[width=\linewidth]{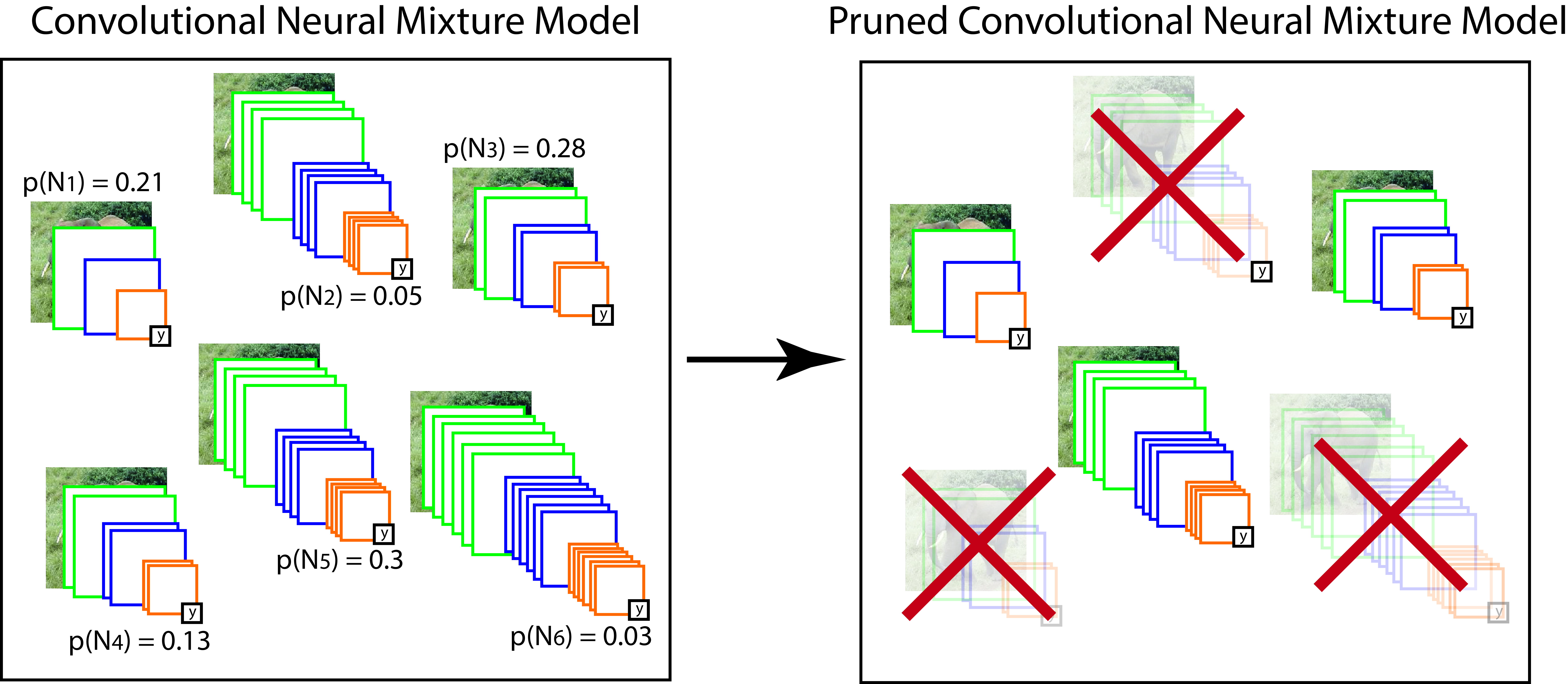}
  \figvspaceOne
  \caption{A Convolutional Neural Mixture Model embeds a large number of CNNs. 
  Weight sharing enables efficient joint training of all networks and computation of the mixture output. 
  The learned mixing weights can be used to remove networks from the mixture, and thus reduce the computational cost  of inference.}
  \figvspaceTwo
  \label{fig:summary}
\end{figure}

To accelerate CNNs we can reduce their complexity before training, \eg by decreasing the number of filters or network layers. 
This solution, however, may lead to sub-optimal results given that over-parametrization plays a critical role in the optimization of deep networks~\cite{du2018gradient,frankle2018lottery}. 
Fortunately, other studies have   found a complementary phenomena: given a trained CNN, a large number of its filters are redundant and do not have a significant impact on the final prediction~\cite{jaderberg2014speeding}. Motivated by these two findings, much research has focused on  accelerating CNNs using network pruning~\cite{ghosh2018structured,huang2018condensenet,huang2018data,li2016pruning,liu2017learning,louizos2017learning,neklyudov2017structured,veniat2018learning}. 
Pruning  can be applied at multiple levels, \eg by removing independent filters~\cite{li2016pruning,louizos2017learning}, groups of them~\cite{ghosh2018structured,huang2018condensenet}, or entire layers~\cite{veniat2018learning}. 
Despite the  encouraging results of these methods, their ability to provide a wide range of operating points along the trade-off between accuracy and computation is limited. 
The reason is that these approaches typically require to train a separate model for each specific pruning level.

In this paper, we propose Convolutional Neural Mixture Models (CNMMs), 
which provide a novel perspective on network pruning.
A CNMM define a distribution over a large number of CNNs. 
The mixture is naturally pruned  by  removing networks with low probabilities, see \fig{summary}. 
Despite the appealing simplicity of this approach, it presents several challenges. 
First, learning  a large ensemble of CNNs may require a prohibitive amount of computation. 
Second, even if many networks in the mixture are pruned, their independent evaluation during inference is likely to be less efficient than computing the output of a single large model.

In order to ensure tractability, we design a  parameter-sharing scheme  between different CNNs. 
This  enables us to (i) jointly train all the networks, and 
(ii) efficiently compute an approximation of the mixture output without independently evaluating all the networks. 

Image classification and semantic segmentation experiments  show that  CNMMs achieve an excellent trade-off between prediction accuracy and computational cost. 
Unlike most  previous network pruning approaches, a single CNMM model achieves a wide range of  operating points  along this trade-off without any re-training.

\section{Related work}

\mypar{Neural network ensembles} Learning ensembles of neural networks is a long-standing research topic. Seminal works explored different strategies to combine the outputs of different networks to obtain more accurate predictions~\cite{krogh1995neural,naftaly1997optimal,zhou2002ensembling}. Recently, the success of deep models has renewed interest in ensemble methods. 

For this purpose, many  approaches have been explored. For instance,~\cite{lakshminarayanan2017simple,zhu2018binary} used  bagging~\cite{breiman1996bagging} and boosting~\cite{schapire2003boosting} to train multiple networks. 
Other works have considered to learn diverse models by employing different parameter initializations~\cite{lee2015m}, or re-training a subset of layers~\cite{zhao2018retraining}. 
While these strategies are effective to learn diverse networks, their main limitation is the required training cost. 
In practice, training a deep model can take multiple days,  and therefore  large  ensembles may have a prohibitive cost. 
To reduce the training time, it has been suggested~\cite{huang2017snapshot,loshchilov2016sgdr} to train a single network and to use parameters from  multiple iterations of the optimization process to define the ensemble. Despite the efficiency of this method during training, this approach does not reduce  inference cost, since   multiple networks must be evaluated independently at test time.

An alternative strategy to allow efficient training and inference is to use implicit ensembles~\cite{ghosh2018structured,huang2016deep,larsson2016fractalnet,neklyudov2017structured}. 
By relying on  sampling, these methods allow to jointly train all the individual components in the ensemble and perform approximate inference during testing. 
Bayesian neural networks (BNNs) fall in this paradigm and use  a distribution over  parameters, rather than a single point estimate~\cite{ghosh2018structured,kingma2015variational,louizos2017multiplicative,neklyudov2017structured}.
A sample from the parameter distribution can be considered as an individual network. Other works have implemented the notion of implicit ensembles by using dropout \cite{srivastava2014dropout} mechanisms. 
Dropping neurons can be regarded as sampling over a large ensemble of different networks~\cite{baldi2013understanding}. Moreover, scaling outputs during testing according to the dropout probability can be understood as an approximated inference mechanism. 
Motivated by this idea, different works have applied dropout over individual weights~\cite{gal2017concrete}, network activations~\cite{singh2016swapout}, or connections in multi-branch architectures~\cite{han2017branchout,larsson2016fractalnet}. 
Interestingly, it has been observed that ResNets~\cite{he2016identity} behave like an ensemble of models, where some residual connections can be removed without significantly reducing prediction accuracy~\cite{veit2016residual}. 
This idea was used by ResNets with stochastic depth~\cite{huang2016deep}, where different dropout probabilities are assigned to the residual connections. 

\begin{figure*}[t]
\includegraphics[width=\textwidth]{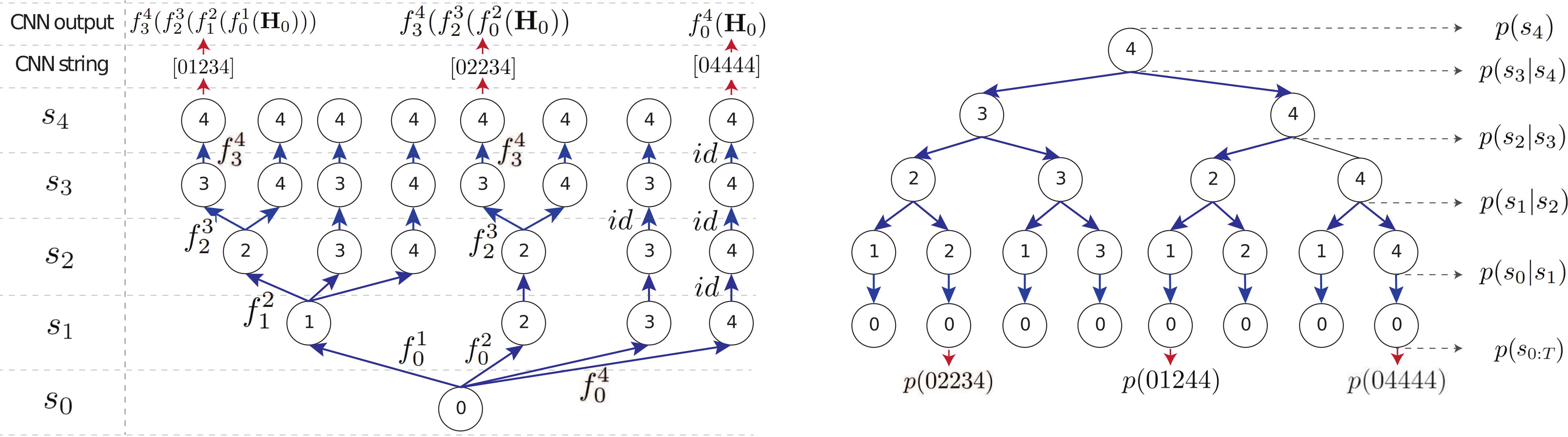}
\figvspaceOne
\caption{(Left) Illustration of how a large collection of CNNs is represented in a CNMM. 
Each network is uniquely identified by a non-decreasing sequence $s_{0:T}$, containing numbers from $0$ to $T$. 
Consecutive entries     in the sequence determine the functions $f_{s_{t-1}}^{s_t}$ applied to compute the CNN output. 
In this manner, sequences with common sub-sequences share functions and their parameters in the corresponding networks.
(Right) Illustration of the distribution $p({s}_{0:T})$ that defines the mixing weights over the CNN models, here  $T=4$. 
Each $p(s_{t-1}|s_t)$ is  a Bernoulli distribution on whether  $s_{t-1}$ equals $s_t$ or $t-1$. 
This defines a binary tree generating all the valid  sequences.}
\label{fig:nmm_illustration}
\figvspaceTwo
\end{figure*}

Our proposed Convolutional Neural Mixture Model is an implicit ensemble defining a mixture distribution over an exponential number of CNNs. This allows to use the learned probabilities to prune the model by removing non-relevant networks. Using a mixture of  CNNs for model pruning is a novel approach, which contrasts to previous methods employing ensembles for other purposes such as boosting performance~\cite{dutt2018coupled,lee2015m}, improving learning dynamics \cite{huang2016deep}, or uncertainty estimation~\cite{ilg2018uncertainty,lakshminarayanan2017simple}.

\mypar{Efficient inference in deep networks} 
A number of strategies  have been developed to reduce the inference time of CNNs, including the design of 
%Many works have attempted to reduce the inference time of CNNs by using several strategies such as designing
efficient convolutional operators~\cite{howard2017mobilenets,iandola2016squeezenet,zhang2018shufflenet}, knowledge distillation~\cite{chen2017learning,hinton2015distilling}, neural architecture search~\cite{he2018amc,zoph2018learning}, weight compression~\cite{masana2017domain,tai2015convolutional}, and quantization~\cite{jacob2018quantization,lin2016fixed}. 
Network pruning has emerged as one of the most effective frameworks for this purpose~\cite{ghosh2018structured,lecun90nips,li2016pruning,veniat2018learning}. 
Pruning methods aim to remove weights which do not have a significant impact on the network output. 
Among these methods, we can differentiate between two main strategies: online and offline pruning. 
In offline pruning, a network is first optimized for a given task using standard training. Subsequently, non-relevant weights are identified using different heuristics including their norm~\cite{li2016pruning},  similarity to other weights~\cite{srinivas2015data}, or second order derivatives~\cite{lecun90nips}. 
The main advantage of this strategy is that it can be applied to any pre-trained network. However, these approaches require a costly process involving several prune/retrain cycles in order to recover the original network performance. Online approaches, on the other hand, perform pruning  during network training. 
For example, sparsity inducing regularization can be used over individual weights~\cite{liu2017learning,louizos2017multiplicative}, groups of them~\cite{ghosh2018structured,huang2018condensenet,neklyudov2017structured}, or over the connections in multi-branch architectures~\cite{ahmed2018maskconnect,veniat2018learning}. 
These methods typically have a hyper-parameter, to be set before training,  determining the trade-off between the final performance and the pruning ratio. 

In contrast to previous approaches, we prune entire CNNs  by removing the networks with the smallest probabilities in the mixture. 
This approach offers two main advantages. 
First, it does not require to define a hyper-parameter before training to determine the balance between the potential compression and the final performance. 
Second, the number of removed networks can be controlled after optimization. 
Therefore, a learned CNMM can be deployed at multiple operating points to trade-off computation and prediction accuracy. 
For example, across different  devices with varying computational resources, or on the same device with different computational constraints depending on the processor load of other processes. The recently proposed Slimmable Neural Networks \cite{yu2018slimmable} have also focused on adapting the accuracy-efficiency trade-off at run time. This is achieved by embedding a small set of CNNs with varying widths into a single model. Different from this approach, our CNMMs embed a large number of networks with different depths, which allows for a finer granularity to control the computational cost during pruning.

%%%%%%%%%%%%%%%%%%%%%%%%%%%%%%%%%%%%%%%%%%%%%%%%%%%%%%%%%%%%%%%%%%%%%%%%%%%%%%%%%%
\section{Convolutional Neural Mixture Models}
%%%%%%%%%%%%%%%%%%%%%%%%%%%%%%%%%%%%%%%%%%%%%%%%%%%%%%%%%%%%%%%%%%%%%%%%%%%%%%%%%%
Without loss of generality, we consider a CNN as a function $\mathcal{F}( \mathbf{H}_0)=\mathbf{H}_T$ mapping an RGB image $ \mathbf{H}_0 \in \mathbb{R}^{W_0 \times H_0 \times 3}$ to a tensor $\mathbf{H}_T \in  \mathbb{R}^{W \times H \times C}$. 
In particular, we assume that $\mathcal{F}$ is defined as a sequence  of $T$  operations:
\begin{equation}
    \mathcal{F}(\mathbf{H}_0) = f_{T-1}^{T}(\dots (f_{1}^2(f_0^{1}(\mathbf{H}_0)))),
    \label{eq:cnn_def}
\end{equation}
where $\mathbf{H}_t=f_{t-1}^t(\mathbf{H}_{t-1})$ is computed from the previous  feature map $\mathbf{H}_{t-1}$.  We assume that the functions $f_{t-1}^t$ can be either the identity function,  or a standard CNN block composed of different operations such as batch-normalization, convolution, activation functions, or spatial pooling. 
In this manner, the effective depth of the network, \ie the number of non-identity layers $f_{t-1}^t$, is at most $T$. 

The output tensor $\mathbf{H}_T$ of the CNN is used to make  predictions for a specific task. 
For example, in image classification, a linear classifier over $\mathbf{H}_T$ can be used in order to estimate the class probabilities for the entire image.
For semantic segmentation the same linear classifier is used for each spatial position in $\mathbf{H}_T$.

Given these definitions, a convolutional neural mixture model (CNMM) defines a distribution over  output $\mathbf{H}_T$  as:
\begin{eqnarray}
    p(\mathbf{H}_T|\mathbf{H}_0) & = & \sum_{\mathcal{F} \in \mathds{F}} p(\mathcal{F}) p(\mathbf{H}_T|\mathbf{H}_0 ,\mathcal{F}),
    %p(\mathbf{H}_L|\mathbf{H}_0,\mathcal{F}) & = & \delta(\mathbf{H}_L-\mathcal{F}(\mathbf{H}_0)),
    \label{eq:nmm_def}
\end{eqnarray}
where $\mathds{F}=\{\mathcal{F}_1,\mathcal{F}_2,...,\mathcal{F}_K\}$ is a finite set of CNNs, $p(\mathbf{H}_T|\mathbf{H}_0,\mathcal{F}_k)$ is a delta function centered on the output $\mathcal{F}_k(\mathbf{H}_0)$ of each network, and $p(\mathcal{F})$ defines the mixing weights over the CNNs in $\mathds{F}$.

%%%%%%%%%%%%%%%%%%%%%%%%%%%%%%%%%%%%%
\subsection{Modelling a distribution over CNNs}
\label{sec:ensemble_definition}
%%%%%%%%%%%%%%%%%%%%%%%%%%%%%%%%%%%%%%%%%%%%%%%%%%%%%%%%

%%%%%%%%%%%%%%%%%%%%%%%%%%%%%%%%%%%%%%%%%%%%%%%%%%%%%%%%%%%%%%%%%%%%%%%%%%%%%%%%%%%%%%%%%%%%%
We now  define mixtures that contain a number of CNNs that is exponential in the maximum depth $T$, in a way that allows us to manipulate these mixtures in a tractable manner. 

Each component in the mixture is a chain-structured CNN  uniquely characterised by a sequence  ${s}_{0:T}$ of length $T+1$, where the sequences are constrained to be  a non-decreasing set of integers from $0$ to $T$,  \ie with   $s_0=0$,  $s_{T}=T$ and $s_{t+1}\geq s_{t}$. 
This sequence determines the set of functions that are used in \Eq{cnn_def}. 
In particular, given a sequence ${s}_{0:T}$, the output of the corresponding network is computed as:
\begin{equation}
    \mathcal{F}(\mathbf{H}_0) = f_{s_{T-1}}^{s_{T}}(\dots (f_{s_{1}}^{s_{2}}(f_{s_0}^{s_1}(\mathbf{\mathbf{H}_0})).
    %\mathcal{F}(\mathbf{h}_0) = f_T(\dots (f_t( f_1(\mathbf{h}_0, \theta_{s_0s_1}),\theta{s_Ts_{T-1}}),\theta{s_Ts_{T-1}}
    \label{eq:cnn_def2}
\end{equation}
For $i<j$ the function $f_i^j$ is a convolutional block as  described above with its own parameters, while the functions $f_i^i$ are  identity functions that leave the input unchanged. 

By, imposing $s_{t-1}\in \{t-1, s_t\}$, there is a one-to-one mapping between sequences ${s}_{0:T}$ and the corresponding CNNs.\footnote{In particular, this constraint ensures that, \eg, the network $f_1^4(f_0^1(\mathbf{H}_0))$ is uniquely encoded by the sequence `01444', ruling out the alternative sequences  `01144' and `01114'. See \fig{nmm_illustration}{ (Left)}. }
If multiple  networks  use the same function $f_i^j$, these networks share their parameters on this function, which ensures that the total number of parameters of the mixture does not grow exponentially, although there are exponentially many mixture components. 
For instance, for $T=4$, the mixture will be composed of eight different networks illustrated in \fig{nmm_illustration} (Left). 
From the illustration it is easy to see that, in general, the mixture contains  $2^{T-1}$ components with shared parameters.

In order to define the probabilities $p(\mathcal{F})$ for each network in the mixture, we define a distribution over sequences ${s}_{0:T}$ as a reversed Markov chain:
\begin{equation}
    \label{eq:seq_distribution_fw}
    p({s}_{0:T}) = p({s}_T) \prod_{t=1}^{T} p({s}_{t-1} | {s}_{t}).
\end{equation}
To ensure that sequences have positive probability if and only if they are valid, \ie satisfy the constraints defined above, we set $p(s_T\!=\!T)=p(s_0\!=\!0 | s_1 )=1$ and define:
\begin{equation}
    \label{eq:seq_distribution_probs_bw}
    p(s_{t-1} | s_t ) = 
    \begin{cases} 
        \pi_{t-1}^{s_t}   & \text{if} \hspace{3mm}  {s}_{t-1} = (t-1), \\
        1-\pi_{t-1}^{s_t} &  \text{if} \hspace{3mm} {s}_{t-1} = {s}_{t}, \\
        0   &  \text{otherwise}.
    \end{cases}
\end{equation}

As illustrated in \fig{nmm_illustration}{ (Right)}, these constraints generate a binary tree generating valid non-decreasing sequences. The conditional probabilities $p(s_{t-1}|s_t)$ are modelled by a Bernoulli  distribution with probability $\pi_{s_{t-1}}^{s_t}$, indicating whether the previous number in the sequence is $s_t$ or $t-1$.

%%%%%%%%%%%%%%%%%%%%%%%%%%%%%%%%%%%%%%%%%%%%%%
\subsection{Sampling outputs from CNMMs}
\label{sec:sampling_nmms}
%%%%%%%%%%%%%%%%%%%%%%%%%%%%%%%%%%%%%%%%%%%%%%%

\begin{figure}[t]
  \centering
  \includegraphics[width=0.85\linewidth]{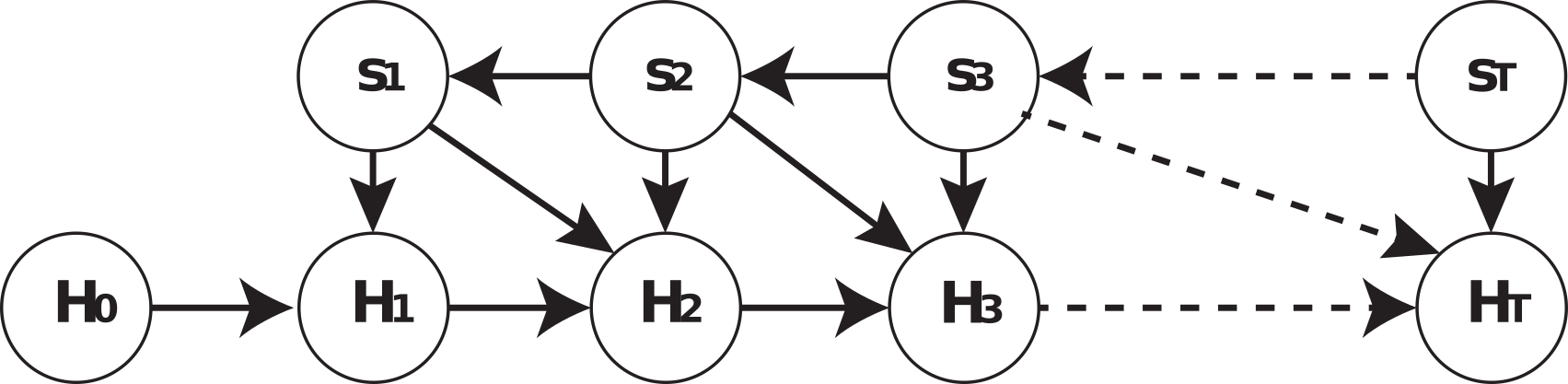}
  %\figvspaceOne
  \caption{Graphical model representation of the CNMM. 
  The sequence $s_{1:T}$ codes for the CNN architecture.
  Each $\mathbf{H}_t$ is an intermediate feature map generated by the sampled CNN. 
  It is computed from the previous feature map $\mathbf{H}_{t-1}$ using  $f_{s_{t-1}}^{s_{t}}$.}
  \label{fig:nmm_pgm}
  \figvspaceTwo
\end{figure}

The graphical model defined in \fig{nmm_pgm}  shows that we  can sample from the output distribution $p(\mathbf{H}_T|\mathbf{H}_0)$ in Eq.~(\ref{eq:nmm_def}) by first generating a sequence from $p({s}_{0:T})$ and then evaluating the associated network with \Eq{cnn_def2}. 
In the following, we formulate an alternative  strategy to  sample from the model. 
This formulation offers two advantages. 
(i) It is amenable to continuous relaxation, which facilitates learning.
(ii) It suggests an iterative algorithm to compute feature map expectations, which can be used instead of sampling for efficient inference. 

The conditional $p(\mathbf{H}_t|s_t=l,\mathbf{H}_0)$ 
gives the distribution over $\mathbf{H}_t$ across  the networks with $s_t=l$. 
For  example,    $p(\mathbf{H}_2|s_2=4,\mathbf{H}_0)$ consists of two weighted delta peaks, located  at  $f_0^4(\mathbf{H}_0)$ and $f_0^1(f_1^4(\mathbf{H}_0))$, respectively. See \fig{nmm_illustration}{ (Left)}.
These conditional distributions can be expressed as the  forwards recurrence:
\begin{align}
    \label{eq:recursive_term}  
    p(\mathbf{H}_t|s_t,\mathbf{H}_0) =  \sum_{\mathclap{\mathbf{H}_{t-1},{s}_{t-1}}} \Big[ p(\mathbf{H}_t |  & \mathbf{H}_{t-1}, {s}_t,{s}_{t-1}) p({s}_{t-1} | {s}_{t}) \nonumber \\ 
    & \times \underbrace{p(\mathbf{H}_{t-1} | {s}_{t-1},\mathbf{H}_0)}_{\text{Recurrent term}} \Big],
\end{align}
where $p(\mathbf{H}_t | \mathbf{H}_{t-1}, {s}_t,{s}_{t-1})$  is a delta function centered on $f_{s_{t-1}}^{s_t}(\mathbf{H}_{t-1})$.
Therefore, unbiased samples  $\tilde{\mathbf{h}}^{s_t}_t$ from $ p(\mathbf{H}_t|s_t,\mathbf{H}_0)$ 
can be obtained through sample propagation. Recall from \Eq{seq_distribution_probs_bw} that, given $s_t$, there are only two possible values of  $s_{t-1}$ that remain, namely  $s_t$ and $t-1$. As a consequence, the sum over $\mathbf{s}_{t-1}$ in Eq. (\ref{eq:recursive_term}) only consists of two terms. Given this observation, samples  $ \tilde{\mathbf{h}}^{s_t}_t\sim p(\mathbf{H}_t|s_t,\mathbf{H}_0)$ can be obtained from  samples  $\tilde{\mathbf{h}}^{s_{t-1}}_{t-1}\sim p(\mathbf{H}_{t-1}|s_{t-1},\mathbf{H}_0)$ as:
\begin{equation}
        \label{eq:sampling_conditional}
        \tilde{\mathbf{h}}^{s_t}_t = 
           \tilde{\pi}_{t-1}^{s_t} f_{t-1}^{s_t}(\tilde{\mathbf{h}}^{t-1}_{t-1})  
        + (1-\tilde{\pi}_{t-1}^{s_t})\tilde{\mathbf{h}}^{s_{t}}_{t-1},
\end{equation}
where for a given value of $s_t$
we sample $s_{t-1}$ from $p(s_{t-1} | s_t)$  to compute a binary indicator $\tilde{\pi}_{t-1}^{s_t}=\Ind{s_{t-1} = t-1}$, which signals whether the resulting $\tilde{\mathbf{h}}^{s_t}_t$ is equal to $\tilde{\mathbf{h}}^{s_t}_{t-1}$ or $f_{t-1}^{s_t}(\tilde{\mathbf{h}}^{t-1}_{t-1})$.

\begin{figure}[t]
  \includegraphics[width=\linewidth]{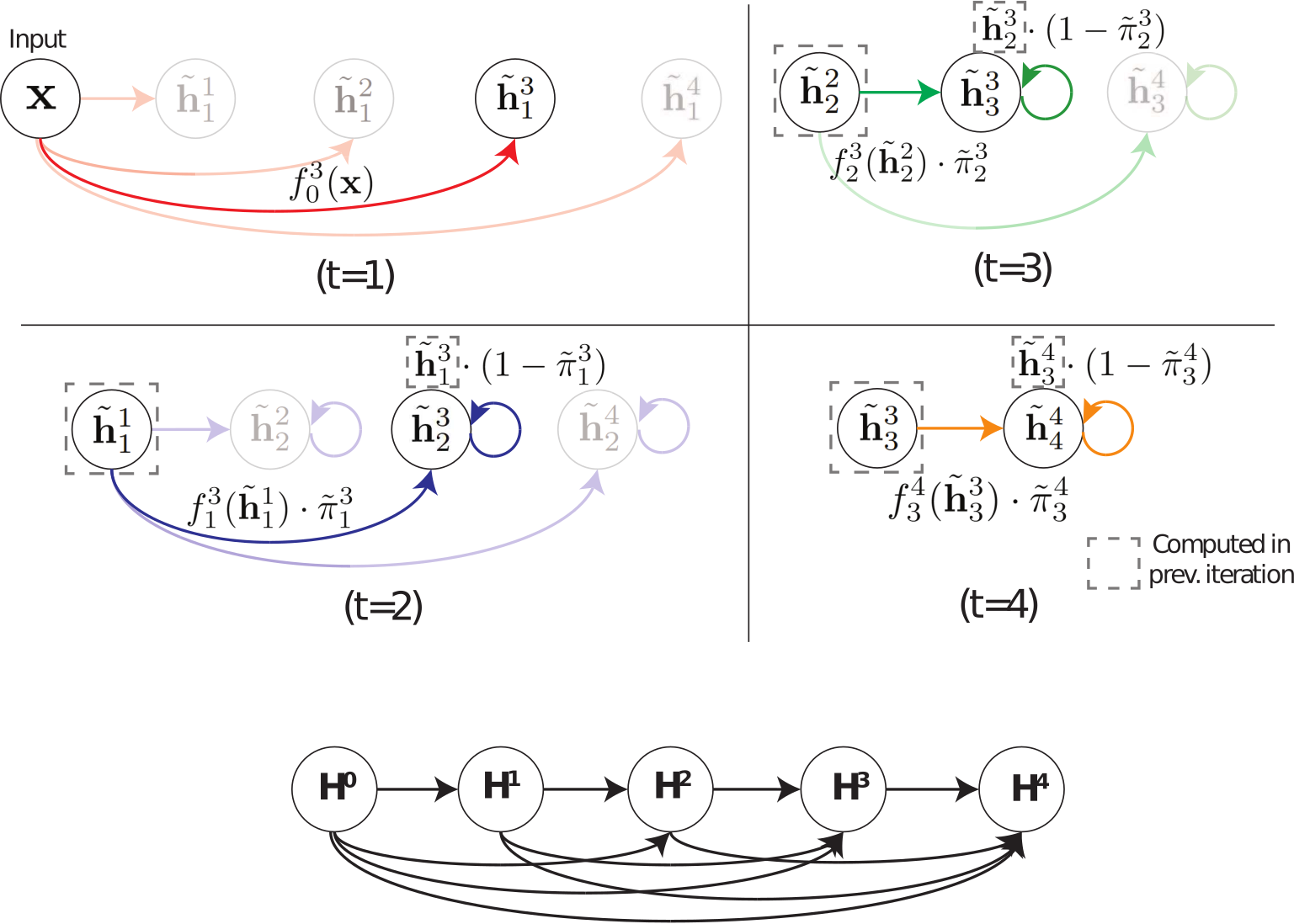}
  \figvspaceOne
  \caption{Top: Illustration of the algorithm used to sample intermediate feature maps from the mixture distribution. At each iteration $t$, we generate $\tilde{\mathbf{h}}^{s_t}_t \sim p(\mathbf{H}_t|s_t,\mathbf{H}_0)$ by using: (i) samples $\tilde{\mathbf{h}}^{t-1}_{t-1}$ obtained in the previous iteration, (ii) the corresponding functions $f_{t-1}^{s_t}$ and (iii) samples $\tilde{\pi}_{t-1}^{s_t}$ from $p(s_{t-1}|s_t)$. Bottom: Network with dense connectivity implementing the sampling algorithm.}
  \label{fig:nmm_inference}
  \figvspaceTwo
\end{figure}

Using \Eq{sampling_conditional} we iterative sample from distributions $p(\mathbf{H}_t|s_t,\mathbf{H}_0)$ for $t=1,\dots,T$, and for each $t$ we compute samples for  $s_t=t,\dots,T$.
An illustration of the algorithm is shown in \fig{nmm_inference}. 
The computational complexity of a complete pass in this iterative process is $O(T(T+1)/2)$, since for each $t=1,\dots,T$, we compute $T-t+1$ samples, each of which is computed in $O(1)$ from the samples already computed for $t-1$.
This is roughly equivalent to the cost of evaluating a single network with dense layer connectivity of depth $T$~\cite{huang2017densely}, which has a total of $T(T-1)/2$ connections implemented by the functions $f_i^j$.

\mypar{Sampling outputs from networks of bounded depth} 
Using the described algorithm, $\tilde{\mathbf{h}}^T_T \sim p(\mathbf{H}_T|s_T=T,\mathbf{H}_0)$ correspond to output tensors $\mathbf{H}_L$ sampled from the mixture defined in Eq.~(\ref{eq:nmm_def}). 
Moreover, for any $t$, samples from  $p(\mathbf{H}_t|s_t=T,\mathbf{H}_0)$ are output feature maps generated by networks with depth bounded by $t$.
For instance, in  \fig{nmm_illustration}, samples $\tilde{\mathbf{h}}^T_{2}$ are generated with one of the networks coded by the sequences $01444$  and $04444$. 

%%%%%%%%%%%%%%%%%%%%%%%%%%%%%%%
%%%%%%%%%%%%%%%%%%%%%%%%%%%%%%%

%%%%%%%%%%%%%%%%%%%%%%%%%%%%%%%%%%%%%%%%%%%%%%%%
\subsection{Training and inference}
\label{sec:training_inference_nmms}
%%%%%%%%%%%%%%%%%%%%%%%%%%%%%%%%%%%%%%%%%%%%%
%\mypar{Learning from CNMMs outputs} 
We use $\psi$ to collectively denote the parameters of the convolutional blocks $f_i^j$ and the parameters $\pi_{s_{t-1}}^{s_t}$  defining the mixing weights via \Eq{seq_distribution_probs_bw}. 
Moreover, the parameters of the classifier that predict the image label(s) from the output tensor $\mathbf{H}_T$ are denoted as $\theta$.
Given a training set $\mathcal{D}=\{(\mathbf{X}_1,y_1),\dots,(\mathbf{X}_N,y_N)\}$ composed of images $\mathbf{X}_n$ and labels $y_n$, we optimize the parameters by minimizing 
\begin{eqnarray}
 \label{eq:loss_function}
\mathcal{L}_{\text{single}}(\psi,\theta)   =    \sum_{n=1}^N  \mathbb{E}_{p({\mathbf{H}}_T | \mathbf{H}_0=\mathbf{X_n};\psi)} \Big[ \mathcal{L}\left(y_n,{\mathbf{H}}_T,\theta \right) \Big],
\end{eqnarray}
where $\mathcal{L}\left(y_n,{\mathbf{H}}_T,\theta \right)$  is the cross-entropy loss comparing the label $y_n$ with the class probabilities computed from  $\mathbf{H}_T$. 
In practice, we replace the expectation over $\mathbf{H}_T$ in each training iteration with samples from $p({\mathbf{H}}_T | \mathbf{H}_0=\mathbf{X_n};\psi)$.

%%%%%%%%%%%%%%%%%%%%%%%%%%%%%%%%%%%%%%%%%%%
\mypar{Learning from subsets of networks} 
As discussed in \sect{sampling_nmms}, 
samples from the distribution $p(\mathbf{H}_t|s_t=T,\mathbf{H}_0)$ correspond to  outputs of CNNs in the mixture with depth at most  $t$.
In order to improve performance of models with  reduced inference time, we explicitly emphasize the loss for such efficient relatively shallow networks. 
Therefore, we sum the above loss function over the outputs sampled from networks of increasing depth:
\begin{align}
 \label{eq:loss_function_anytime}
 \mathcal{L}(\theta,\psi)\!=\!\sum_{n=1}^N  \sum_{t=1}^T \mathbb{E}_{p({\mathbf{H}}_t | s_t=T,\mathbf{X_n};\psi)} \Big[ \mathcal{L}\big(y_n,{\mathbf{H}}_t,\theta_t \big) \Big],
\end{align}
where we use a separate classifier for each $t$. In practice, we balance each loss with a weight increasing linearly with $t$.

\mypar{Relaxed binary variables with concrete distributions} 
The recurrence in \Eq{sampling_conditional} requires  sampling  from  $p(s_{t-1}|s_t)$, defined in \Eq{seq_distribution_probs_bw}. 
The sampling  renders the parameters $\pi_{t-1}^{s_t}$  non-differentiable, which  prevents gradient-based optimization for them. 
To address this limitation, we use a continuous relaxation by modelling $p(s_{t-1}|s_t)$ as a binary ``concrete'' distribution~\cite{maddison2016concrete}. In this manner, we can use the re-parametrization trick~\cite{kingma14iclr,rezende14icml}  to back-propagate gradients \wrt samples $\tilde{\pi}_{t-1}^{s_t}$ in \Eq{sampling_conditional} and, thus to compute gradients for the parameters $\pi_{t-1}^{s_t}$.

\mypar{Efficient inference by expectation propagation} 
Once the CNMM is trained, the predictive distribution on $y$ is given by $p(y|\mathbf{X};\theta) = \mathbb{E}_{p(\mathbf{H}_T|\mathbf{X})}[p(y|\mathbf{H}_T;\theta)]$.
The expectation is intractable to compute exactly, contrary to our goal of efficient inference.
A naive Monte-Carlo sample approximation is still  requires multiple evaluations of the full CNMM. Instead, we propose an alternative approximation by propagating expectations instead of samples  in \Eq{sampling_conditional}, \ie using the approximation 
$\bar{\mathbf{H}}_T \approx p(\mathbf{H}_T|\mathbf{X})$, where $\bar{\mathbf{H}}_T$ is obtained by running the sampling algorithm replacing the samples $\tilde{\pi}_{t-1}^{s_t}$ with their expectations ${\pi}_{t-1}^{s_t}$.

%%%%%%%%%%%%%%%%%%%%%%%%%%%%%%%%%%%%%%%%%%%%%%
\subsection{Accelerating CNNMs}
\label{sec:pruning}
%%%%%%%%%%%%%%%%%%%%%%%%%%%%%%%%%%%%%%%%%%%%%%%
CNMMs offer two complementary mechanisms in order to accelerate inference. We describe both in the following.

\mypar{Evaluating intermediate classifiers} 
The different classifiers $\theta_t$ learned by minimizing \Eq{loss_function_anytime} operate over the outputs of a mixture of  networks with maximum depth $t$. 
Therefore, at each iteration $t$ of the inference algorithm in \Eq{sampling_conditional} we can already output predictions based on classifier $\theta_t$.
This strategy is related with the one employed in multi-scale dense networks (MSDNets)~\cite{huang2017multi}, where ``early-exit'' classifiers are used to provide  predictions at various points in time  during the inference process. 

\mypar{Network pruning} 
A complementary strategy to accelerate CNMMs is to remove  networks from the mixture. 
The computational cost of the  inference process  is dominated by the evaluation of the CNN blocks   $f_{t-1}^{s_t}(\tilde{\mathbf{h}}^{t-1}_{t-1})$ in \Eq{sampling_conditional}.
However, these function does not need to be computed when the variable $\tilde{\pi}_{t-1}^{s_t}=0$. Therefore, a natural approach to prune CNMMs is to set  certain ${\pi}_{t-1}^{s_t}$ to zero, 
 removing  all the CNNs from the mixture that use $f^{s_t}_{t-1}$.  
We  use the learned distribution $p({s}_{0:T})$ in order to remove networks with a low probability. 
Note that for a given value of $s_t$, the pairwise marginal $p(s_t, s_{t-1}\!=\!t-1)$ is exactly the sum of probabilities of all the networks involving the function $f^{s_t}_{{t-1}}$. Using this observation, we use an iterative pruning algorithm where, at each step, we compute all pairwise marginals $p(s_t,s_{t-1}\!=\!t-1)$ for all possible values of $s_t$ and $t$. 
We then set  ${\pi}_{t^\star-1}^{s_t^\star}=0$ where
$(s_t^\star$,$t^\star) = \arg\min_{(s_t,t)} p(s_t,s_{t-1}\!=\!t-1)$.
Finally, the marginals are updated, and we iterate. 

%%%%%%%%%%%%%%%%%%%%%%%%%%%%%%%
\begin{figure}[t]
\centering
  \figvspaceOne
  \includegraphics[width=0.9\linewidth]{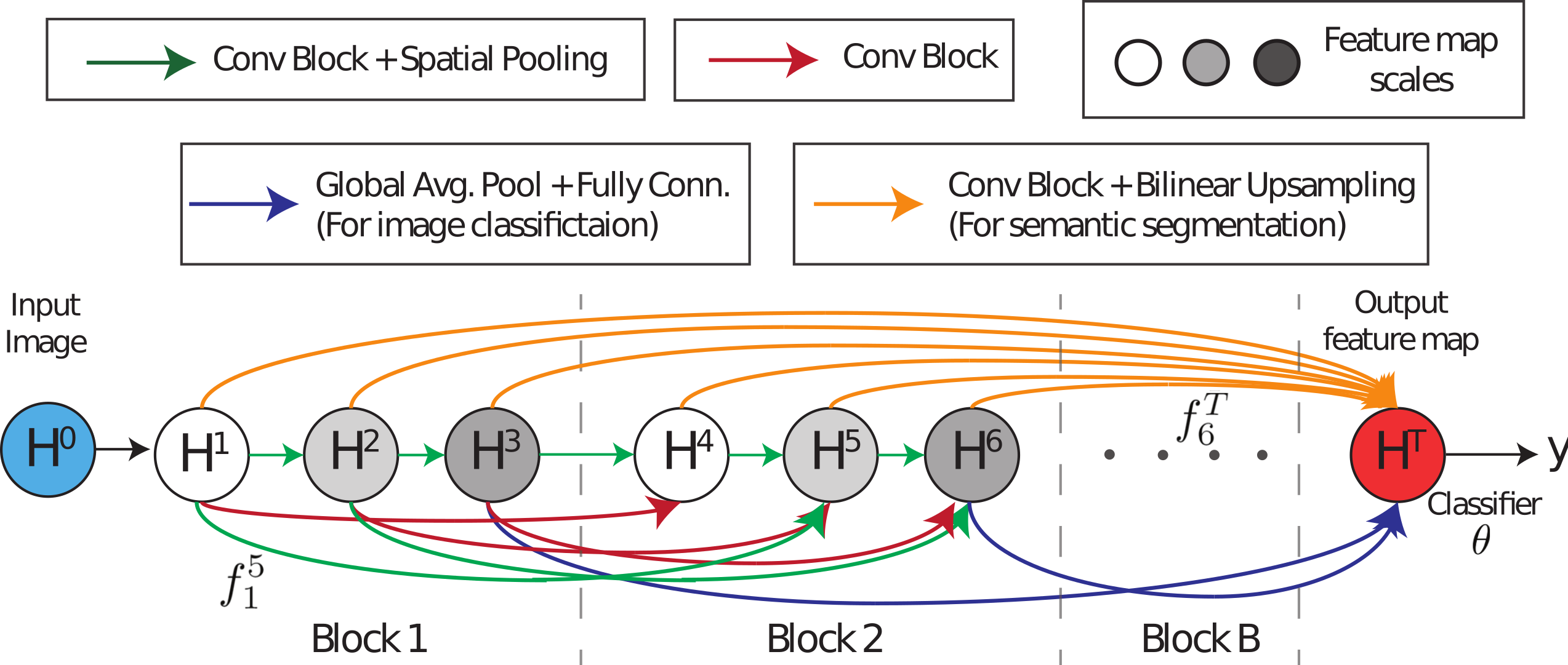}
  \caption{Dense network with sparse connectivity implementing the inference algorithm of our CNMMs. See text for details.}
  \label{fig:cnmm_implementation}
  \figvspaceTwo
\end{figure}
%%%%%%%%%%%%%%%%%%%%%%%%%%%%%%%

%%%%%%%%%%%%%%%%%%%%%%%%%%%%
\begin{figure*}
%\centering
  \includegraphics[width=1\linewidth]{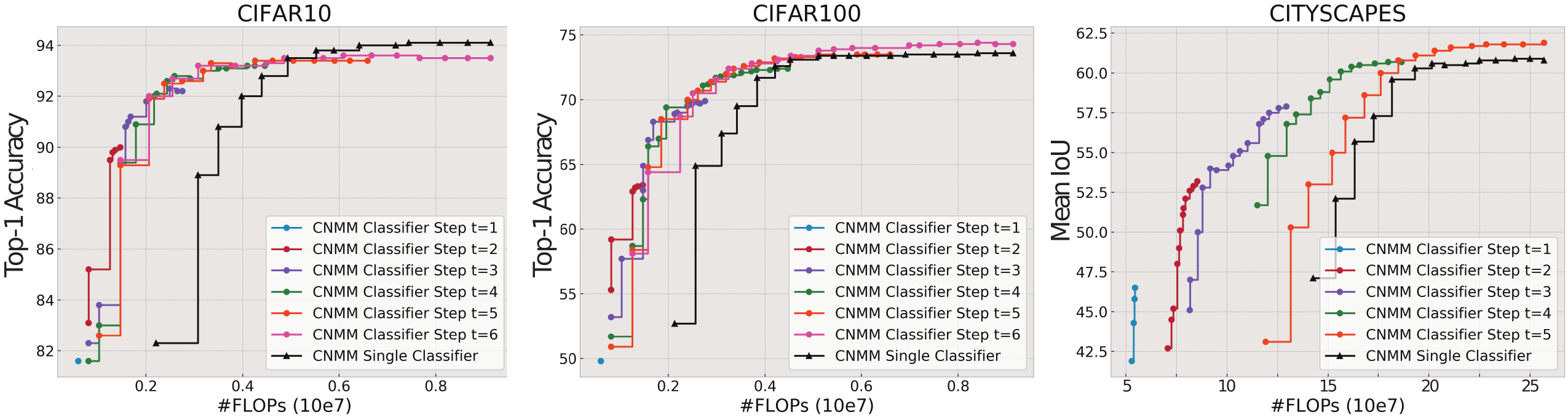}
  \figvspaceOne
  \caption{Prediction accuracy \vs FLOPs for accelerated CNMMs. 
  Black curves depict the performance of a CNMM learned using a single final classifier.
  Colored curves correspond to intermediate classifiers at different steps of the inference algorithm.  
  Points on one curve are obtained by progressively pruning convolutional layers.}
  \figvspaceTwo
  \label{fig:pruning_evaluation}
\end{figure*}
%%%%%%%%%%%%%%%%%%%%%%%%%%%%

In this manner, we  achieve different pruning levels by progressively removing convolutional blocks that will not be evaluated during inference.  This process does not require any re-training of the model, allowing to dynamically set different pruning ratios. Note that this process is complementary to the use of intermediate classifiers, as discussed above. The reason for this is that our pruning strategy may be used to remove   functions $f_i^j$ for any ``early'' prediction  step $t<T$. Finally, it is interesting observe that the proposed pruning mechanism can be regarded as a form of neural architecture search \cite{liu2018darts,zoph2018learning}, where the optimal network connectivity for a given pruning ratio is automatically discovered by taking into account the learned probabilities $p(s_{t-1}\!=\!t-1|s_t)$.

\section{Experiments}
\label{sec:experiments}
We perform experiments over two different tasks: image classification and semantic segmentation. 
Following previous work, we measure the computational cost in terms of the number of floating point multiply and addition operations (FLOPs) required for inference. 
The number of FLOPs provides a metric that correlates very well with the actual inference wall-time, 
while being independent of  implementation and hardware  used for evaluation.

\subsection{Datasets and experimental setup}
\label{sec:datasets}
\mypar{CIFAR-10/100 datasets} These datasets~\cite{krizhevsky2009learning} are composed of 50k train and 10k test images with a resolution of 32$\times$32 pixels. 
The goal is to classify each image across 10 or 100 classes, respectively. 
Images are normalized using the means and standard deviations of the RGB channels. 
We apply standard data augmentation operations: (i) a $4$-pixel zeros padding followed by 32$\times$32 cropping. (ii) Random horizontal flipping with probability $0.5$. 
Performance is evaluated in terms of the mean accuracy across classes.

\mypar{CityScapes dataset} This dataset~\cite{cordts2016cityscapes} contains 1024$\times$2048 pixel images of urban scenes with  pixel-level labels across 19 classes. 
The dataset is split into training, validation and test sets with 2,975, 500 and 1,525 samples each. 
The ground-truth annotations for the test set are not public, and we use  the validation set instead  for evaluation. 
To assess performance we use the standard mean intersection-over-union (mIoU) metric.
We follow the setup of \cite{mehta2018espnetv2}, and  down-sample the images  by a factor two before processing them. 
As a data augmentation strategy during training, we apply random horizontal flipping and resizing by using a scaling factor between 0.75 and 1.1. 
Finally, we use random crops of 384$\times$768 pixels from the down-sampled images.

%%%%%%%%%%%%%%%%%%%%%%%%%%%%
\begin{figure*}[t]
  \includegraphics[width=\linewidth]{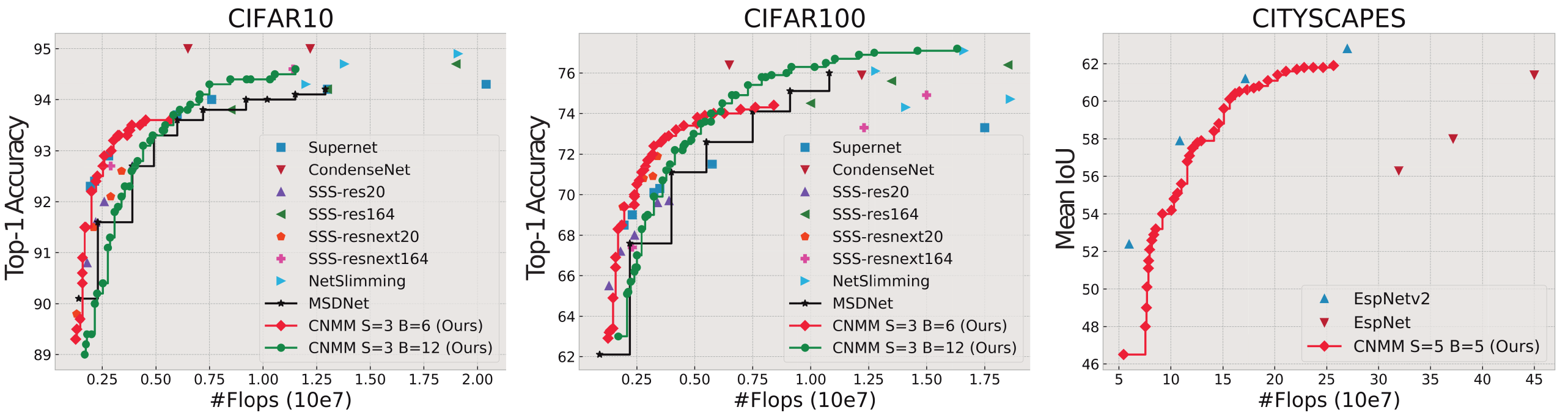}
  \figvspaceOne
  \caption{Comparison of the our CNMM with state-of-the-art  efficient inference approaches  on the  CIFAR and CityScapes datasets. 
  Disconnected markers refer to models that are trained  independently. 
  Curves correspond to a single model that can operate at different number of FLOPs. 
   CNMM curves are obtained by using the optimal combination of pruning and intermediate classifiers.}
  \label{fig:soa_comparison}
  \figvspaceTwo
\end{figure*}
%%%%%%%%%%%%%%%%%%%%%%%%%%%%%%%

\mypar{Base architecture}
As discussed in \sect{sampling_nmms}, the learning and inference algorithms for CNMMs can be implemented  using a  network with dense layer connectivity \cite{huang2017densely}. 
Based on this observation,  we use an architecture similar to MSDNets~\cite{huang2017multi}. Specifically. we define a set of $B$ blocks, each  composed of a set of $S$ feature maps $\mathbf{H}_t$. See Fig. (\ref{fig:cnmm_implementation}).

The initial feature map in each block has $C$ channels and, at each subsequent feature map in the block, the spatial resolution is reduced by a factor two in each dimension,  and the number of channels is doubled. 
Feature maps are connected by functions $f_i^j$ if the output feature map $\mathbf{H}_j$ has the same or half the resolution of the input feature map  $\mathbf{H}_i$.
Finally, we consider the output tensor $\mathbf{H}_T$ to have different connectivity and spatial resolution depending on the task. 

\mypar{Implementation for image classification} 
We implement the convolutional layers as the set of operations (BN-ReLU-DConv-BN-ReLU-Conv-BN), where BN refers to batch normalization, DConv is a $(3\times 3 \times C \times \frac{C}{4})$ depth-wise separable convolution~\cite{howard2017mobilenets}, and Conv is a $(1 \times 1 \times \frac{C}{4} \times C)$ convolution. 
In order to reduce computation, for a given tensor $\mathbf{H}_i$, 
the different functions $f_i^j$ share the parameters of the initial operations (BN-ReLU-DConv) for all $j$. 
Moreover, when the resolution of the feature map is reduced, we use average pooling after these three initial operations. 
In all our experiments, the number of initial channels in $\mathbf{H}_1$ is set to $C=64$. 
This is achieved by using a $(3 \times 3 \times 3 \times \frac{C}{4})$  convolution over the input image $\mathbf{H}_0$ and then apply a $(1 \times 1 \times \frac{C}{4} \times C)$ convolutional block. 
Finally, all the tensors $\mathbf{H}_l$ with the lowest spatial resolution are connected to the output $\mathbf{H}_L$. Concretely, $\mathbf{H}_T$ is a vector $\mathbb{R}^{512}$ obtained by applying the operations (BN-ReLU-GP-FC-BN) to the input tensors, where GP refers to global average pooling, and FC corresponds to a fully-connected layer. The classifier $\theta$ maps  $\mathbf{H}_T$ linearly to a vector of dimension  equal to the number of classes. 
When using \Eq{loss_function_anytime} to train the CNMM, we connect a  classifier $\theta_t$ with the end of each block. 

\mypar{Implementation for semantic segmentation} 
We use the same setup as for image classification, but replace the ReLU activations with parametric ReLUs as in~\cite{mehta2018espnet}. 
Moreover, we use max instead of average pooling  to reduce the spatial resolution. The input tensor $\mathbf{H}_1$ has $C=48$ channels and a resolution four times lower than the original images. 
This is achieved by applying a $(3 \times 3 \times 3 \times \frac{C}{4})$ convolution with stride $2$ to the input and then using a (BN-ReLU-Conv) block followed by max pooling. The output tensor $\mathbf{H}_L$ receives connections from all the previous feature maps and has the same channels and spatial resolution as $\mathbf{H}_1$. 
Given that the input feature maps are at different scales, we apply a (BN-PReLU-Conv-BN) block over the input tensor and use bi-linear up-sampling with different scaling factors in order to recover the original spatial resolution.  The final classifier $\theta$ computing the class probabilities using $\mathbf{H}_T$ are defined as blocks (UP-BN-PReLU-Conv-UP-BN-PReLU-DConv), where UP refers to bilinear upsampling, which allows to recover the original image resolution. The first and second convolutions in the block have $12$ and $K$ output channels respectively, where $K$ is the number of classes. As in image classification, we use an intermediate classifier $\theta_t$ at each step $t$ where a full block of computation is finished.

\mypar{Optimization details} 
In  our experiments, we use SGD with momentum by setting the initial learning rate to 0.1 and weight decay to $10^{-4}$. In CIFAR, we use a cosine annealing schedule to accelerate convergence. On the other hand, in Cityscapes we employ a cyclic schedule with warm restarts as in \cite{mehta2018espnetv2}.
The temperature of the concrete distributions modelling $p(s_{t-1}|s_t)$ is set to 2. 
We train our model by using 300 and 200 epochs, and batch size of 64 and 6, for respectively CIFAR-10/100 and Cityscapes. For the cyclic scheduler, the learning rate is divided by two at epochs $\{30,60,90,120,150,170,190\}$. Additionally, the models trained in Cityscapes are fine-tuned during 10 epochs by using random crops of size 512$\times$1024 instead of 384$\times$768.

%%%%%%%%%%%%%%%%%%%%%%%%%%%%%%%
\begin{figure*}[t]
  \centering
  \includegraphics[width=1.0\linewidth]{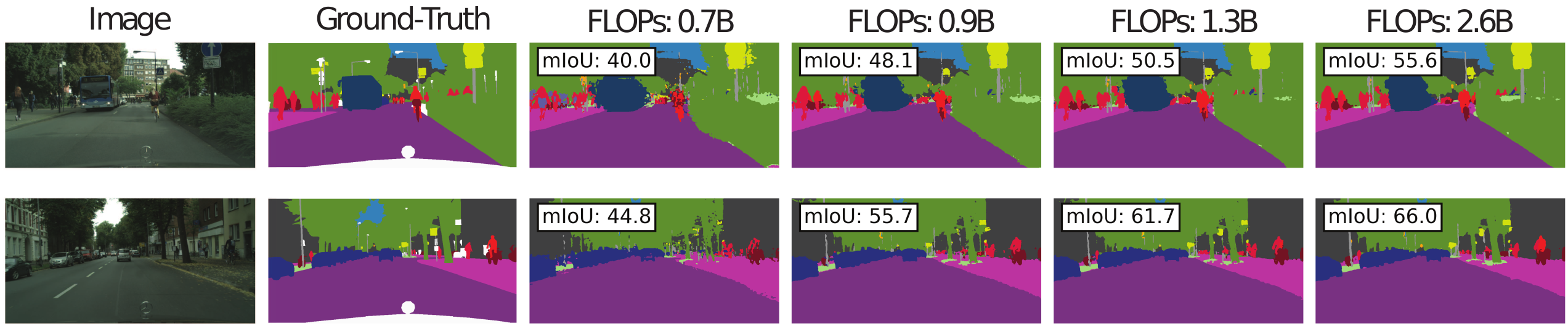}
  \figvspaceOne
  \caption{Pixel-level predictions for a single CNMM operating under different computational constraints. As discussed, our model allows to dynamically dynamically set the trade-off between accuracy and inference time with no additional cost.}
  \label{fig:qual_results}
  \figvspaceTwo
\end{figure*}
%%%%%%%%%%%%%%%%%%%%%%%%%%%%%%%

\subsection{Pruning and intermediate classifiers}
\label{sec:classifiers_eval}

We evaluate the proposed pruning and intermediate classifiers strategies to reduce the inference time of trained CNMMs.
For CIFAR-10/100 we learn a CNMM with $B=6$ blocks, using $S=3$ scales each. 
For Cityscapes we use $B=5$ blocks and $S=5$ scales.
For each dataset,  we train one model that uses  a single classifier $\theta$, optimized  using  \Eq{loss_function}. 
In addition, we train a second model with intermediate classifiers $\theta_t$,  minimizing the loss function in \Eq{loss_function_anytime}. 
In the following, we will refer to the first and second variant as CNMM-single and CNMM respectively.

In \fig{pruning_evaluation} we report  prediction accuracy \vs FLOPs for inference.
Each model is represented as a curve, traced by pruning the model to various degrees. 
Across the three datasets, the CNMM model with intermediate classifiers achieves higher accuracy in fast inference settings than the CNMM-single model.
Recall that  all the operation point across the different CNMM curves are obtained from a single trained model. 
Therefore, this single model can realize the upper-envelope of the performance curves. As expected, the maximum performance of the  intermediate  classifiers increases with the step number. The accuracy of CNMM at the final step is comparable to the level obtained by the CNMM-single model: slightly worse on CIFAR-10, and slightly better at CIFAR-100 and CityScapes.
This is because the minimized intermediate losses provide additional supervisory signal which is particularly useful to encourage accurate prediction for  shallow, but fast, CNNs.
In conclusion, the CNMM model with intemediate classifiers is to be preferred, since it provides a better trade-off between accuracy and computation at  a wider range of FLOP counts.

By analysing the  operating points along each curve, we can observe the effectiveness of the proposed pruning algorithm. 
For the  CIFAR datasets we can reduce the FLOP count by a factor two  without significant loss in accuracy. 
For CityScapes, about 25\% pruning can be achieved without a significant loss. % due to the higher difficulty of the task. 
In general, if several exit points can achieve the same FLOP count by applying varying amounts of pruning, 
best performance is obtained pruning less for an earlier classifier, rather than pruning more for a later exit.

%%%%%%%%%%%%%%%%%%%%%%%%%%%%%%%%%%%%%%%%%%%%%%%%%%
\subsection{Comparison with the state of the art}
\label{sec:comparison_soa}
%%%%%%%%%%%%%%%%%%%%%%%%%%%%%%%%%%%%%%%%%%%%%%%%%%

\mypar{Image classification} 
We compare our model with different state-of-the-art CNN  acceleration strategies~ \cite{huang2017multi,huang2018condensenet,huang2018data,liu2017learning,veniat2018learning}. 
We consider methods applying pruning at different levels, such as independent filters (Network slimming~\cite{liu2017learning}), groups of weights (CondenseNet)~\cite{huang2018condensenet}, connections in multi-branch architectures  (SuperNet)~\cite{veniat2018learning}, or a combination of them (SSS)~\cite{huang2018data}.  
We also compare our method with any-time models based on early-exit classifiers (MSDNet)~\cite{huang2017multi}. Among other previous state-of-the-art methods, the compared approaches have shown the best performance among efficient inference methods with $\leq$200 million FLOPs.
We compare to CNMMs using 6 and 12 blocks, using three scales is  both cases.

The results in \fig{soa_comparison} (left, center) show that CNMMs achieve similar or better accuracy-compute trade-off  across a broad range of  FLOP counts  than all the compared methods in the CIFAR datasets. 
Only CondenseNets shows somewhat better performance for medium FLOP counts. 
Moreover, note that the different operating points shown
for the compared methods (except for MSDNets) are obtained by using different models trained independently, \eg by different settings of a hyper-parameter controlling the pruning ratio. In contrast, CNMM embeds a large number operating points in a single model. 
This feature is interesting when the available computational budget  can change dynamically, based on concurrent processes, or when the model is deployed across a wide range of devices. 
In these scenarios, a single CNMM can be accelerated on-the-fly depending on the available resources. 
Note that a single MSDNet is also able to provide early-predictions by using intermediate classifiers. 
However, our CNMM provides better performance for a given FLOP count and  allows for a finer granularity to control the computational cost.

\mypar{Semantic segmentation} 
State-of-the-art methods for real-time semantic segmentation have mainly focused on the manual-design of efficient network architectures. 
By employing highly optimized convolutional modules, ESPNet~\cite{mehta2018espnet} and ESPNetv2~\cite{mehta2018espnetv2} have  achieved  impressive accuracy-computation trade-offs. 
Other  methods, such as~\cite{chen2018deeplab,zhao2017pyramid}, offer higher accuracy  but at several orders of magnitude higher inference cost, limiting their application in resource constrained scenarios.

In \fig{soa_comparison} (right) we compare our  CNMM results to these two approaches. Note that the original results reported in \cite{mehta2018espnetv2} are obtained by using a model pre-trained in ImageNet. For a fair comparison with our CNMMs, we have trained EspNetv2 from scratch by using the code provided by the authors \footnote{https://github.com/sacmehta/EdgeNets}.
As can be observed, CNMM provides a better trade-off compared to ESPNet. 
In particular, a full CNMM without pruning obtains an improvement of 0.5 points of mIoU, while reducing the FLOP count by 45\%. 
Moreover, an accelerated CNMM achieves a similar performance compared to the most efficient ESPNet that needs more than two times more FLOPs. 
On the other hand, ESPNetv2 gives slightly better trade-offs compared to our CNMMs. 
However, this model relies on 
an efficient inception-like module~\cite{szegedy2016rethinking} that also includes group point-wise and  dilated convolutions. These are orthogonal  design choices that can be integrated in our model as well, and we expect that to further improve our results.
Additionally, the different operating points in ESPNet and ESPNetv2 are achieved using different models trained independently.
Therefore, unlike our approach, these methods do not allow for a fine-grained control over the accuracy-computation trade-off, and multiple models need to be trained. 
\fig{qual_results} shows qualitative results using different operating points from a single CNMM.

\section{Conclusions}
We proposed to address model pruning by using Convolutional Neural Mixture Models (CNMMs), a novel probabilistic framework that embeds a mixture of an exponential number of CNNs. In order to make training and inference tractable, we rely on massive parameter sharing across the models, and use concrete distributions to differentiate across the discrete sampling of mixture components. 
To achieve efficient inference in CNMM we use an early-exit mechanism that allows prediction after evaluating only a subset of the networks. In addition, we use a pruning algorithm to remove CNNs that have low mixing probabilities. 
Our experiments on image classification and semantic segmentation tasks show that CNMMs achieve excellent trade-offs between prediction accuracy and computational cost.
Unlike most of previous works, a single CNMM model allows for a large number and wide range of accuracy-compute trade-offs, without any re-training. 

\subsection*{Acknowledgements}
This work is supported by ANR  grants ANR-16-CE23-0006 and ANR-11-LABX-0025-01. 

{\small
\bibliographystyle{ieee_fullname}
\bibliography{adria}
}
\clearpage
\begin{appendices}
%%%%%%%%%%%%%%%%%%%%%%%%%%%%

\section{Supplementary material}

We provide further results on CIFAR100 in order to show the importance of all components of our proposed CNMMs. Moreover, we provide additional qualitative results of semantic segmentation on the CityScapes dataset.

\subsection{Ablative study of CNMM}
\label{sec:ablation_studies}

\begin{figure}[t]
  \centering
  \includegraphics[width=\linewidth]{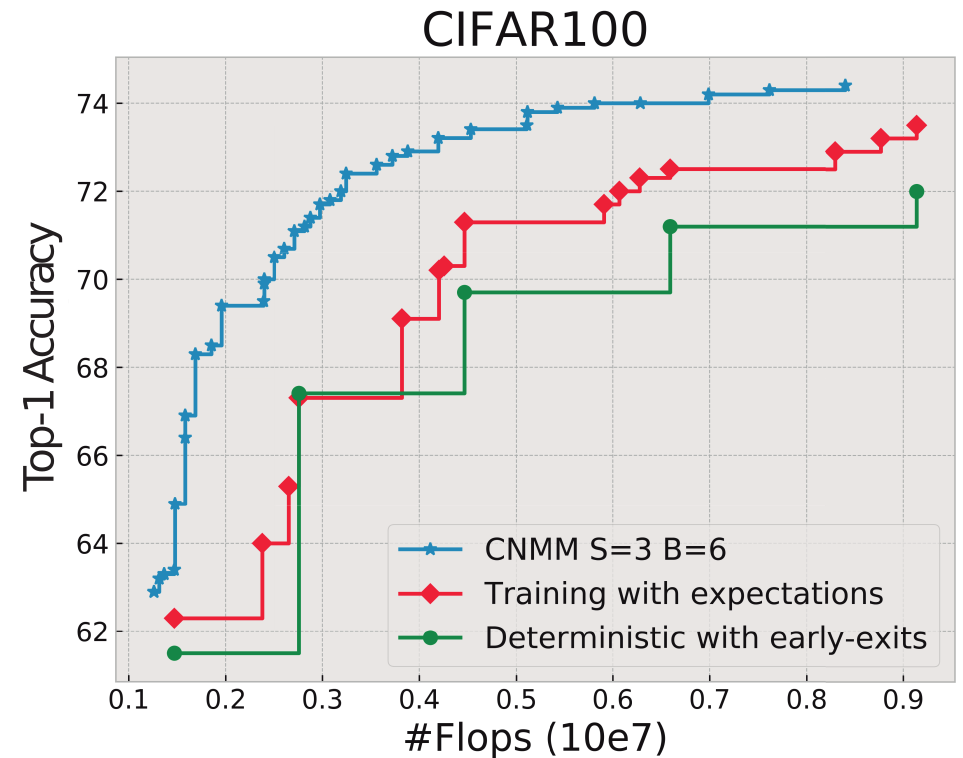}
  \caption{FLOPs \vs performance curves for our model (CNMM), and the  variants described in  \sect{ablation_studies}. 
  As in \fig{soa_comparison} of the main paper, the curves are obtained by using the optimal combination of early-exits and pruning when possible. In this manner, results for CNMM represent upper-envelope of all the different curves depicted in Fig. \ref{fig:pruning_evaluation} of the main paper.}
  \label{fig:ablation_results}
\end{figure}

\mypar{Using sampling during training} 
During learning, CNMMs generate a set of samples $\tilde{\pi}^{s_t}_{{t-1}}$ using  \Eq{sampling_conditional}. 
In contrast, during inference we use the expectations $\pi^{s_t}_{{t-1}}$ instead. 
In order to evaluate the importance of sampling during learning, we have optimized a CNMM by using the aforementioned expectations instead of samples. 
\fig{ablation_results} shows the results obtained by the model using this approach, denoted as ``Training with expectations''. 
We observe that, compared to the CNMM using sampling, the accuracy decreases faster when different pruning ratios are applied. We attribute this to the fact that our sampling procedure can be regarded as a continuous-relaxation of dropout, where a subset of functions $f_{t-1}^{s_t}(\mathbf{h}_{t-1}^{t-1})$ are randomly removed when computing the output tensor $\mathbf{h}_{t-1}^{s_t}$. 
As a consequence, the learned model is more robust to the pruning process where some of the convolutional blocks are removed during inference. 
This is not the case when deterministic expectations are used in \Eq{sampling_conditional} rather than samples.

\mypar{Comparison with a deterministic model} 
We compare the performance of our CNMM with a deterministic variant using the same architecture. 
Concretely, in \Eq{sampling_conditional} we ignore samples $\tilde{\pi}^{s_t}_{{t-1}}$  and simply sum the feature maps $\tilde{\mathbf{h}}^{s_{t}}_{t-1}$ and $\tilde{\mathbf{h}}^{t-1}_{t-1}$. 
Note that the resulting model is analogous to a MSDNet~\cite{huang2017multi} using early-exit classifiers. 
We report the results in \fig{ablation_results}, denoted as ``Deterministic with early-exits''. 
We observe that our CNMM model obtains better performance than its deterministic counterpart. 
Moreover, same as MSDNets, accelerating the deterministic model is only possible by using the early-exits. In contrast, the complementary pruning algorithm available in CNMM allows for a finer granularity to control the computational cost.

\mypar{Expectation approximation during inference}
In order to validate our approximation of $p(y|\mathbf{X};\theta) = \mathbb{E}_{p(\mathbf{H}_T|\mathbf{X})}[p(y|\mathbf{H}_T;\theta)]$ during inference, we evaluate the performance obtained by using a Monte-Carlo procedure for the same purpose. 
In particular, we generate $N$ samples from the output distribution $p(\mathbf{H}_T|\mathbf{H}_0)$.  
Then, we compute the class probabilities $p(y|\mathbf{H}_T;\theta)$ for each sample and average them.
\tab{montecarlo} shows the results obtained by varying the number of samples. 
We observe that  our approach offers a similar performance as the Monte-Carlo approximation using $N=5$. 
For a higher number of samples, we observe slight improvements in the results. However, note that a Monte-Carlo approximation is very inefficient  since it requires $N$ independent evaluations of the model. 

In particular, the last row in \tab{montecarlo} is 30 times more costly to obtain than the two first rows.
The minimal gain obtained with more samples could probably be more efficiently obtained by using a larger model.

%%%%%%%%%%%%%%%%%%%%%%%%%%%%%%%
%%%%%%%%%%%%%%%%%%%%%%%%%%%%%%%+

\begin{figure*}[t]
  \includegraphics[width=\linewidth]{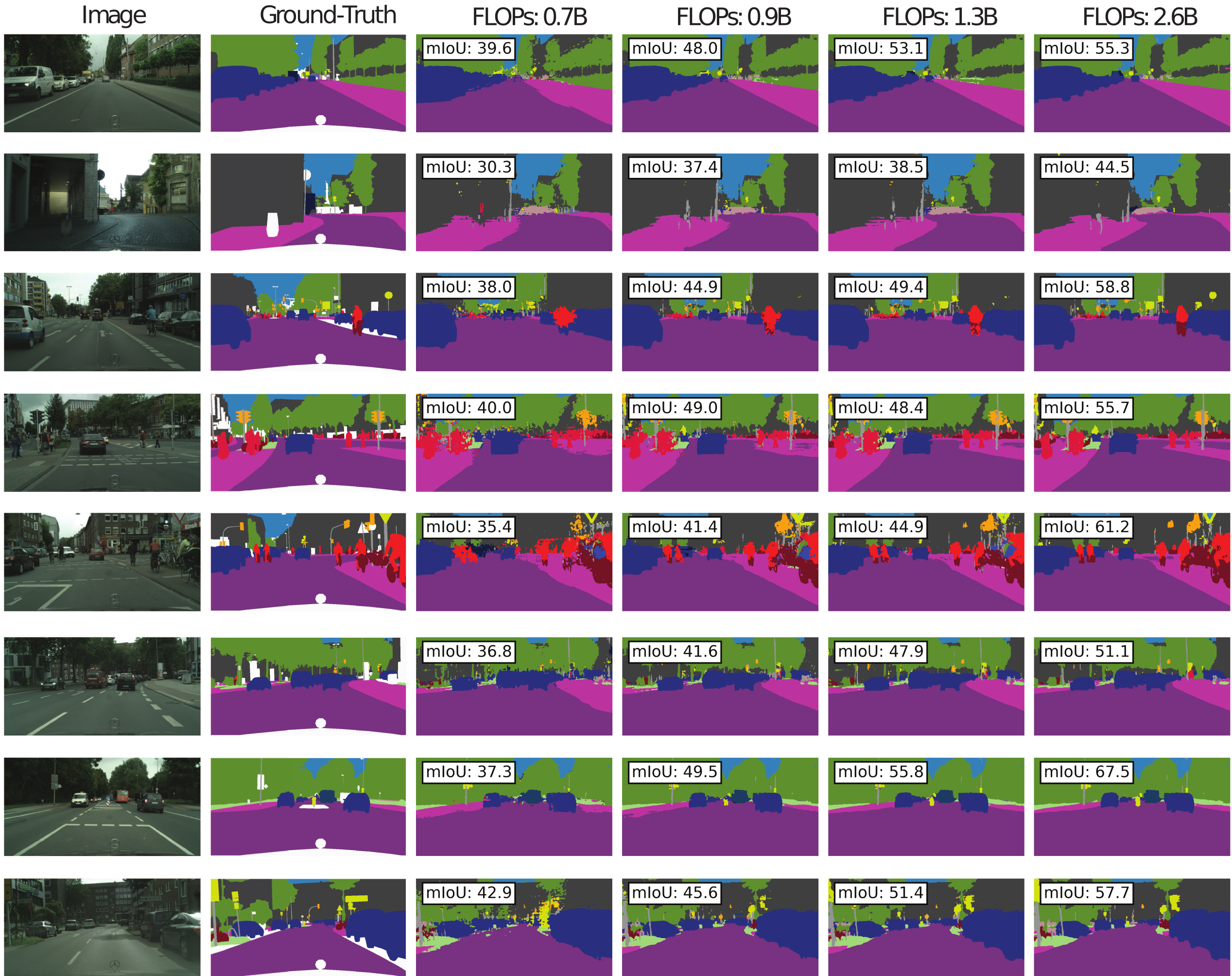}
  \figvspaceOne
  \caption{Pixel-level predictions for a single CNMM adapting the number of FLOPs required during inference.}
  \label{fig:qual_results_supp}
  \figvspaceTwo
\end{figure*}

% \usepackage{graphicx}
\begin{table}[h]
\centering
\resizebox{0.75\linewidth}{!}{%
\begin{tabular}{c|c|c|c|c}
\textbf{Approximation} & \textbf{FLOPs} & \textbf{Top-1 Accuracy}  \\ \hline
\textbf{Expectation (used)} & 93M & 74.4 \\ \hline
\textbf{Sampling N=1} & 93M  & 71.2  \\ \hline
\textbf{Sampling N=5} & 463M  & 74.4 \\ \hline
\textbf{Sampling N=15} & 1390M  & 74.5  \\ \hline
\textbf{Sampling N=30} & 2780M & 74.6  \\ \hline
\end{tabular}%
}
\caption{Comparison of the results obtained in CIFAR100 by approximating the CNMM output using our approach or a Monte-Carlo procedure with different number of samples $N$.}
\label{tab:montecarlo}
\end{table}

\subsection{Additional Qualitative Results}
In \fig{qual_results_supp} we provide additional qualitative results for semantic segmentation obtained by a single trained CNMM model, using various opertating points with different number of FLOPs during inference. 
\end{appendices}
\end{document}